\documentclass[letterpaper, 10 pt, conference]{ieeeconf} 

\IEEEoverridecommandlockouts
\usepackage{cite}
\usepackage{amsmath,amssymb,amsfonts}
\usepackage{algorithmic}
\usepackage{graphicx}
\usepackage{textcomp}
\usepackage{xcolor}
\usepackage{csvsimple}

\newcount\Comments  
\usepackage{color}
\definecolor{darkgreen}{rgb}{0,0.5,0}
\definecolor{purple}{rgb}{1,0,1}
\definecolor{teal}{rgb}{0,0.4627,0.5804}
\newcommand{\kibitz}[2]{\ifnum\Comments=1\textcolor{#1}{#2}\fi}

\def\BibTeX{{\rm B\kern-.05em{\sc i\kern-.025em b}\kern-.08em
    T\kern-.1667em\lower.7ex\hbox{E}\kern-.125emX}}
\begin{document}
\title{SAILing CAVs: Speed-Adaptive Infrastructure-Linked Connected \& Automated Vehicles\thanks{This work was supported by NSF Grants 2135579 and 2111688.}
\thanks{$^{1}$ Institute for Software Integrated Systems, Vanderbilt University}
}

\author{ Matthew Nice$^1$, Matthew Bunting$^1$, George Gunter$^1$, William Barbour$^1$, Jonathan Sprinkle$^1$, Dan Work$^1$}

\maketitle

\begin{abstract}
This work demonstrates a new capability in roadway control: Speed-adaptive, infrastructure-linked connected and automated vehicles. We develop and deploy a lightly modified vehicle that is able to dynamically adjust the vehicle speed in response to posted variable speed limit messages generated by the infrastructure using LTE connectivity. This work describes the open source hardware and software platform that enables integration between infrastructure-based variable posted speed limits, and existing vehicle platforms for automated control. The vehicle is deployed in heavy morning traffic on I-24 in Nashville, TN. The control vehicle follows the posted variable speed limits, resulting in as much as a 25\% reduction in speed variability compared to a human-piloted vehicle in the same traffic stream.


\end{abstract}


\section{Introduction}

\begin{figure}
    \centering
    \includegraphics[width=\linewidth]{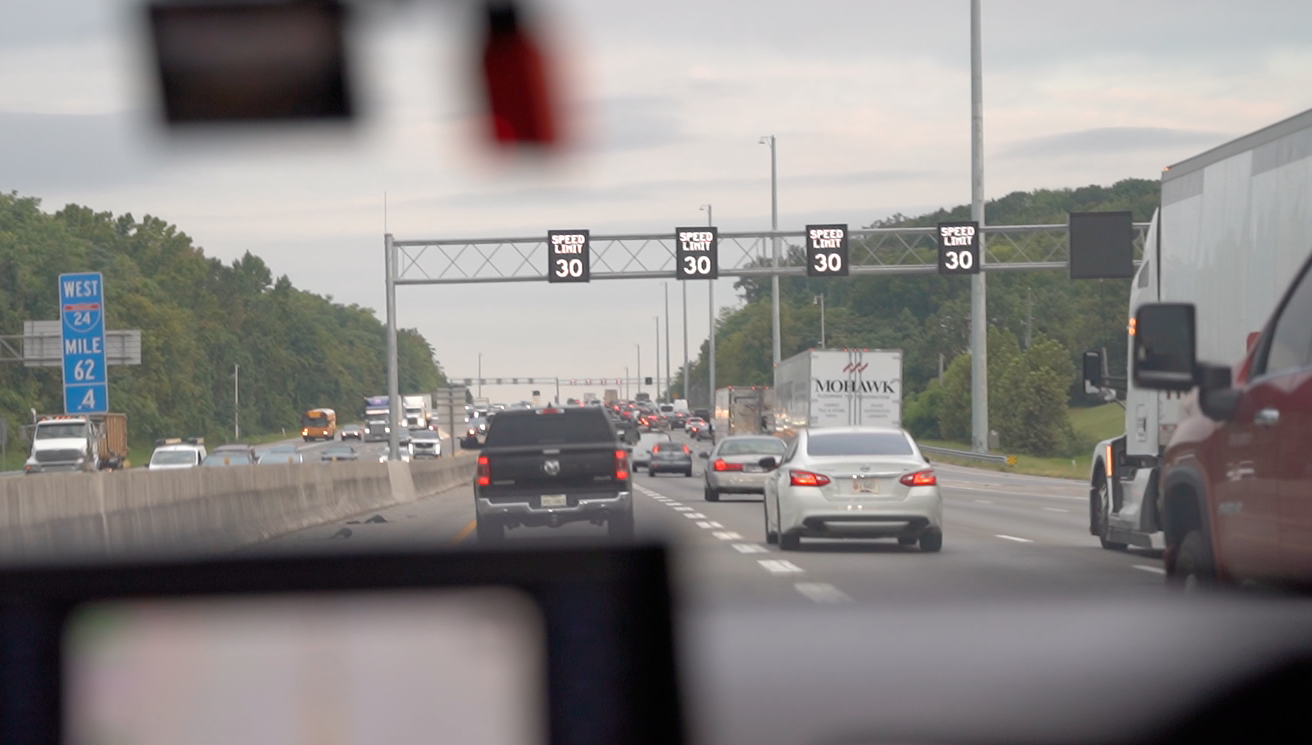}
\caption{Variable speed limit (VSL) system activated and posting 30~mph speed limit during congested traffic. Photo taken from the ego vehicle which automatically adjusts its velocity to 30~mph in response to the VSL system.}
    \label{fig:vsl_active}
\end{figure}

This article introduces a novel capability for \textit{connected automated vehicles} (CAVs) in freeway settings: dynamically adjusting a vehicle's \textit{adaptive cruise control} (ACC) system to adhere to an infrastructure-based variable speed limit system using LTE communication. We deploy this capability on a vehicle in heavy traffic on Interstate 24 near Nashville, TN, which has a new active traffic management system including variable speed limits (Figure~\ref{fig:vsl_active}).

Infrastructure-based \textit{variable speed limit} (VSL) systems have been deployed in several locations around the world over the past few decades~\cite{lu2014review,sergeVSL,papageorgiou2008effects}. These systems have the potential to increase public safety and mobility by adjusting the maximum speed limit to help smooth traffic flow.

One limitation of infrastructure-based VSL is that it requires drivers to comply with the posted speeds to maximize the benefits, which does not always occur in real deployments. Consequently, researchers have been interested in using a small fraction of automated vehicles to improve compliance of the overall traffic stream~\cite{ma2016freeway}, and thereby increasing the benefits~\cite{chen2020effects,han2017variable,khondaker2015variable,shladover2012impacts,yu2019optimal,li2016reducing}. Such works have largely been limited to simulation, due to the lack of widely available CAVs and communication gaps between the infrastructure operators the vehicle automaton systems.

\begin{figure}
    \centering
    \includegraphics[width=\linewidth]{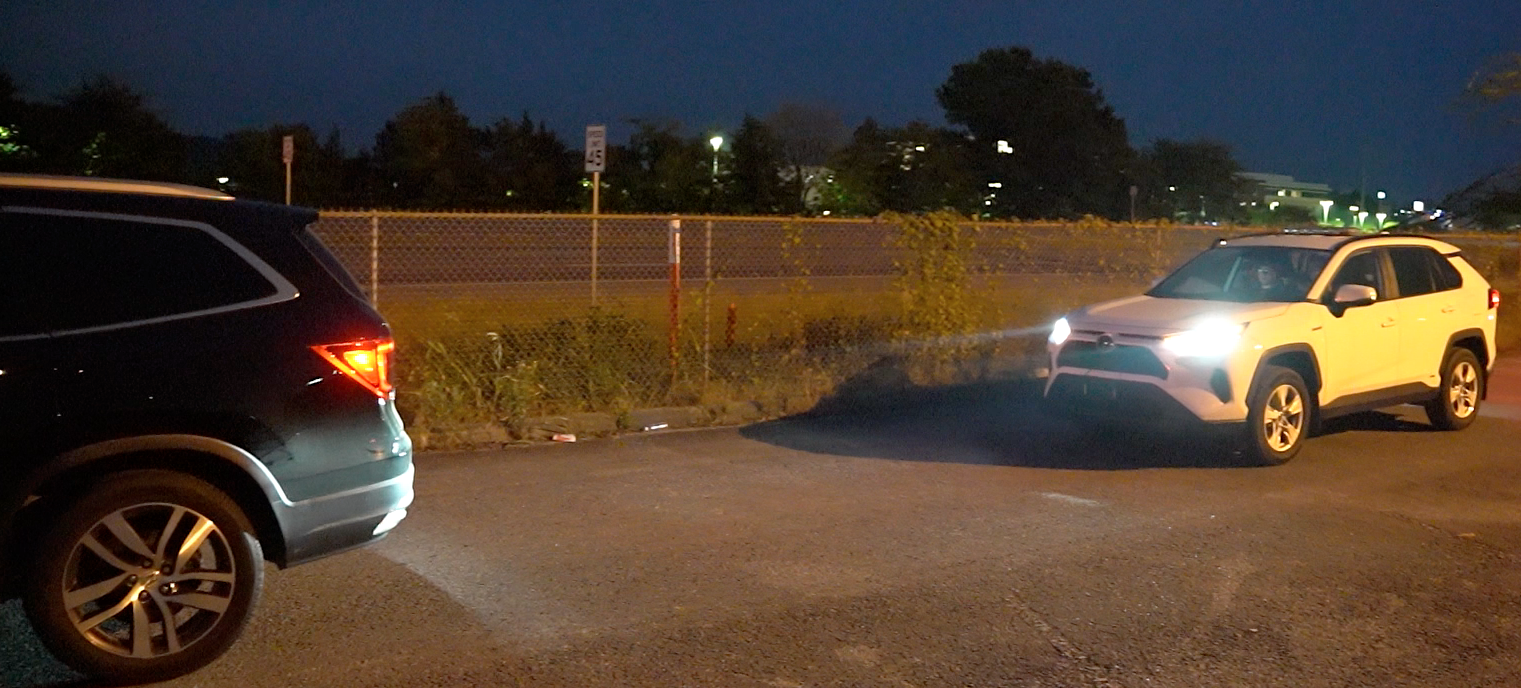}
    \caption{Control vehicle (right) preparing to deploy as a VSL-following CAV on I-24 behind pilot vehicle (left) for trajectory comparison.}
    \label{fig:vu1}
\end{figure}

The main contribution of our work is the development and field deployment of the first connected automated vehicle with the capability to follow publicly broadcast variable speed limits. We implement this capability with open source hardware and software that extends a stock vehicle's adaptive cruise control. We demonstrate the system on a vehicle (Figure~\ref{fig:vu1}) in heavy traffic on an open roadway, and compare the performance of the equipped vehicle to a human piloted vehicle driving in the same traffic.


Our work is enabled by the creation of a low-cost and scalable information pipeline from real public infrastructure into the automated driving of a vehicle. Our implemented system leverages the existing paradigm of the vehicle manufacturer's ACC. The desired speed is set by the driver on all roadways except the whitelisted area for the VSL system, wherein the desired speed is set by the VSL system. The nominal longitudinal control will match the vehicle's speed to the desired speed, unless intervention comes from the supervisory safety controller to facilitate safe car-following and stopping.


Our work builds on the work of~\cite{ma2016freeway}, in which comparably high-cost CAVs are deployed on a congested freeway and adjust their desired speed based on information from 5 fixed roadside sensors. In contrast, our work deploys a low-cost CAV in tandem with active traffic management infrastructure posting a legally enforceable speed limit that is visible to all drivers.  Beyond speed harmonization, connected vehicle field experimentation has occurred in application domains ranging from vehicle platooning\cite{chang1991experimentation,milanes2013cooperative}, to truck safety and efficiency\cite{alan2023integrating}, and demonstrating dedicated short-range communications\cite{bai2006reliability,jerbi2007experimental}.  




The remainder of this article is organized as follows. Section~\ref{sec:methods} describes the Interstate 24 (I-24) infrastructure and the software architectures which extend a stock vehicle into a VSL-following CAV; Section~\ref{sec:results} characterizes the performance of our CAV deployment; and Section~\ref{sec:conclusions} discusses future directions extending from this work.

\section{Methods}\label{sec:methods}

\subsection{Active Traffic Management Infrastructure}
The Interstate-24 SMART Corridor~\cite{tdot2023i24} is an \textit{active traffic management} (ATM) system designed to improve safety and improve travel time variability, particularly during congestion caused by incidents that cause non-recurring congestion. The corridor consists of 28 miles of Interstate 24 (I-24) running in the east/west direction between Nashville and Murfeesboro, Tennessee, USA, along with a parallel section of State Route 1 (SR-1) between the two cities and connector roads between the Interstate and State Route. 

The primary ATM strategies deployed to I-24 are variable speed limits  and lane control systems, in the form of overhead gantries spanning the roadway and spaced every 0.5 miles. 
VSLs are posted at intervals of 1-minute in a range from 30 to 70 mph. Speed limits are chosen with the goal of reducing traffic speed variance, either through a tapered speed reduction algorithm, or through a human operator. The reduction of speed variance through the posted VSLs is intended to improve overall traffic safety. Speed limits are posted on LED message boards above each lane. Radar detection units measure the speed of traffic and are installed on average every 0.35 miles and report speed, occupancy, and volume measurements per lane aggregated over 30-second intervals. These measurements are used by the SMART Corridor traffic operations team in the calculation of posted VSLs.

Notably, not all vehicles need to exhibit strict compliance with the posted speed limit for the system to have a positive effect. A small number of vehicles complying with the speed limit has a greater \textit{effective compliance rate} since non-complying vehicles have limited ability to maneuver around complying ones~\cite{zhang2022quantifying}.


\subsection{Vehicle System Architecture}
\begin{figure}
    \centering
    \includegraphics[width=\linewidth]{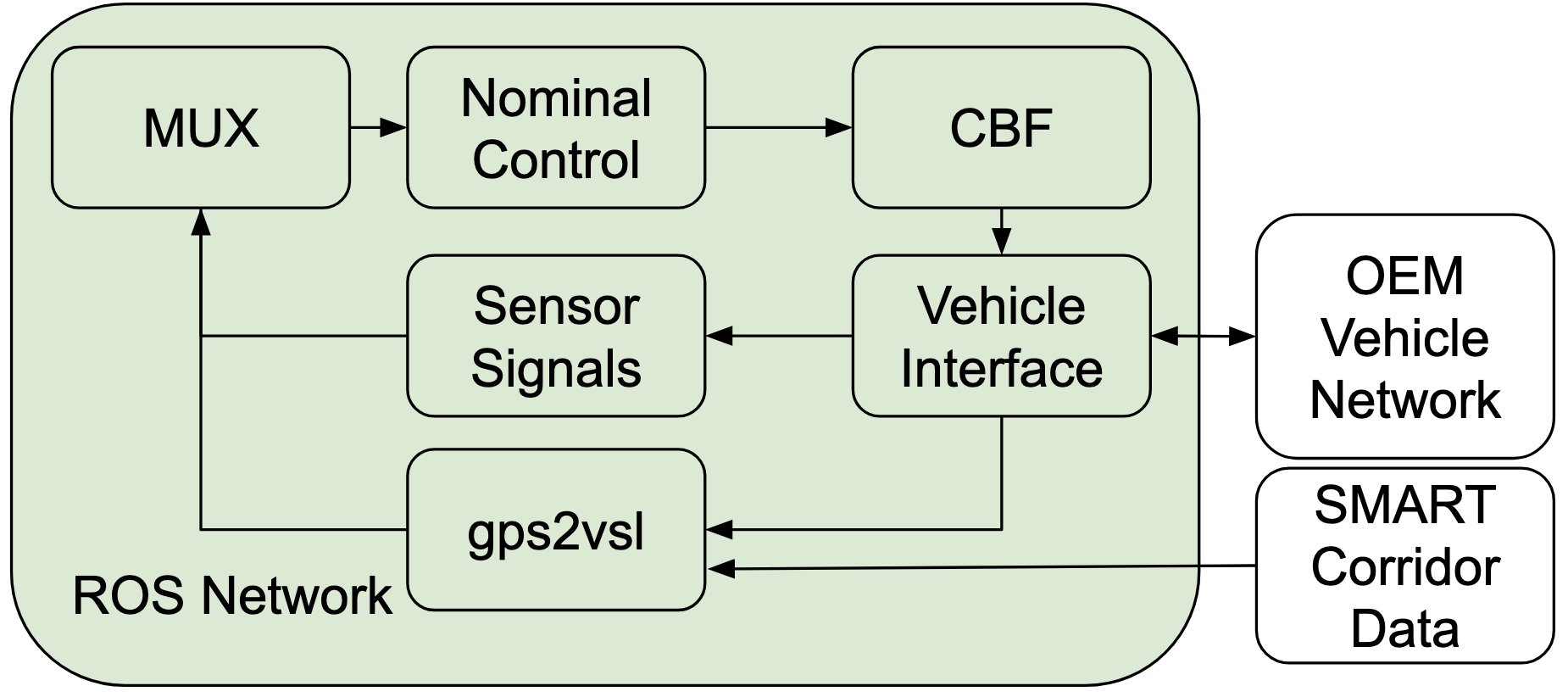}
    \caption{Three networks interact in the VSL-compliant CAV: SMART Corridor (fixed sensor network), OEM Vehicle Network (e.g. radar sensor, drive-by-wire), and the ROS message passing network. Vehicle proprioception, speed recommendations, and safety control actions are handled here.}
    \label{fig:ros}
\end{figure}

Our control vehicle system is implemented on a 2020 Toyota Rav4 pictured in Figure~\ref{fig:vu1}.  Figure~\ref{fig:ros} outlines the vehicle's software system implementation.  Using low cost hardware, the vehicle system accesses the SMART Corridor data through a web-based pipeline over an LTE connection with a tethered mobile phone. Using this information, along with vehicle state propioception, vehicle control commands are created. We leverage the ROS message framework \cite{quigley2009ros} for system design.  Our system can be broken down into three categories: vehicle interfacing, VSL integration, and control design.

\subsubsection{Vehicle Interfaces}

Interfacing with the control vehicle was performed using the software libpanda\cite{bunting2021libpanda}.  Libpanda has the capability to firewall \textit{controller area network} (CAN) messages between system modules designed by the \textit{original equipment manufacturer} (OEM), allowing third party massages to replace OEM messages. Through libpanda, both on-board measurements can be read/recorded, and control commands can be sent to the vehicle. In tandem with with CAN interfacing, libpanda also interfaces with USB GPS modules to provide position information. Libpanda also keeps track of the OEM Advanced Driver Assistance System (ADAS) module state to prevent hardware-level errors when attempting to engage the system.  


Code generation techniques as in \cite{nice2023middleware} are used to convert manufacturer-specific vehicle CAN message into a homogenenous framework in ROS.  In Figure~\ref{fig:ros}, the vehicle interface represents a ROS node with an autogenerated CAN parser that produces sensor data like radar signals and cruise control setpoint. The vehicle interface node is a part of the can\_to\_ros project \cite{elmadani2021can}, exposing CAN-level vehicle systems to ROS.

Multiple setpoints, which are the desired maximum vehicle speeds, are expected to operate the vehicle based on the state of its pose with respect to the I-24 SMART Corridor.  Orchestration of multiple speed setpoint sources is done with a combination of a multiplexer and a ramp function.  This system has two sources of velocity setpoints.  The first is the setting defined by the user as \verb|/user_set_point|.  The second is the posted speed limit provided by the VSL as \verb|/vsl_set_point|.  These can be selected based on the logic defined by the gps2vsl node for when the VSL setpoint is valid (see Section~\ref{sub:gps2vsl}). 

Whether setpoints are changed from multiplexing logic or from updates from the VSL, discrete jumps occur often on the order of $\approx 5\frac{m}{s}$ causing potential transient issues in velocity controllers.  Such issues result in large commanded accelerations or decelerations leading to unsafe behaviors.  To prevent this, a time-based ramp function is placed on the output of the multiplexer to smoothly transition the setpoint. The ramp effectively limits the rate of change the setpoint; in our implementation the setpoint was limited to $1.5\frac{m}{s}$ up and $2\frac{m}{s}$ down. 

\begin{figure}
    \centering
    \includegraphics[width=\linewidth]{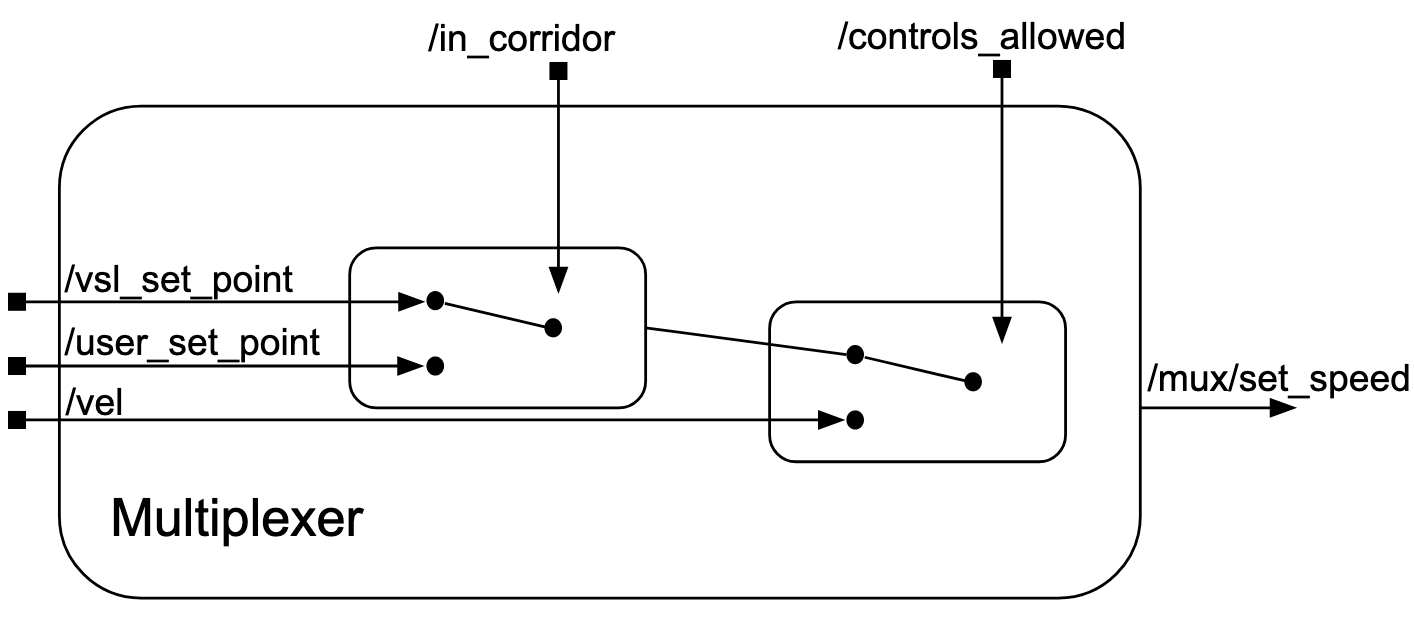}
    \caption{The multiplexer decides which signals are used for the desired velocity using two switches. If libpanda\cite{bunting2021libpanda} is not allowing control, then the desired speed is output as the velocity signal; this feature avoids discontinuities in desired velocities in entry/exit states. If libpanda is allowing control, then the input from a second switch is passed through. In the second switch, if the vehicle is inside the SMART Corridor then the speed recommended by the VSL is the desired speed; otherwise (i.e. if you are anywhere else in the world), the desired speed is set to the set point on the driver dashboard controlled by steering wheel buttons.}
    \label{fig:mux}
\end{figure}

The ramp works well for when the system is engaged however the driver also has the ability to disengage the system, resulting in the vehicle greatly speed mismatching the setpoint.  To prevent transients on system re-engagement, the vehicle's current measured speed from the CAN is provided as \verb|/vel|, and is selected whenever libpanda reports that the system is disengaged.  This lets the ramp closely follow the current speed, preventing memory issues that could lead to large setpoint jumps. Figure~\ref{fig:mux} shows the full structure of the multiplexer.

\subsection{Data Integration from Public Infrastructure}

Messages are sent from the traffic operations center to each gantry whenever the speed limit should change. Our VSL data feed is obtained from a database mirror that records these messages from the traffic operations center to the gantries. 

\subsubsection{Web-based service to establish speed at each gantry}
SMART Corridor data are made available through a three stage process: (1) mirroring of gantry updates into a database; (2) periodic server-side join of default data and gantry updates into a snapshot of VSL system data; and (3) on-demand web-based fetch of the most recent VSL system data.

A database mirror records all changes that are made by traffic operations to the VSL gantries. For efficiency at runtime, our system considers only changes made in the previous 24 hours when establishing the current speed limit. Thus it is necessary for the VSL system web pipeline to include both the default speed limit 
at this stretch of the roadway, and the most recent VSL posted speed 
if it has been updated within the past $T=24$ hours.

The VSL system data are assembled every $15s$ and cached on the server for quick response when requested, rather than executing a query on request. Each row includes the default maximum speed limit for each gantry (in case there have been no traffic operation modifications in the previous window, $T$), and includes whether or not the VSL gantry is ``triggered'' meaning it is posting a speed limit lower than the maximum based on decisions by the SMART Corridor ATM system. 

The on-demand web-based infrastructure is a single URL that returns the entire dataframe for use by the car. Although future work may determine that it is more feasible to return a subset of the dataframe, the current uncompressed size of $\approx 15 kB$ means that the return sizes are negligible for a demonstration project. 





\subsubsection{Models for gps2vsl}
\label{sub:gps2vsl}
The models created for the automated VSL application take inputs from the GPS module, on-board vehicle CAN data, and the VSL system to output a useful desired velocity, or `set speed', to the vehicle's experimental control algorithms. Broadly, this implementation results in a set-and-hold behavior by gantry as the CAV passes VSL gantries travelling through the SMART corridor. Desired speeds are fetched repeatedly to keep the vehicle's velocity tied closely to the posted speed limit as it changes in by location and time.

\begin{figure}
    \centering
    \includegraphics[width=\linewidth]{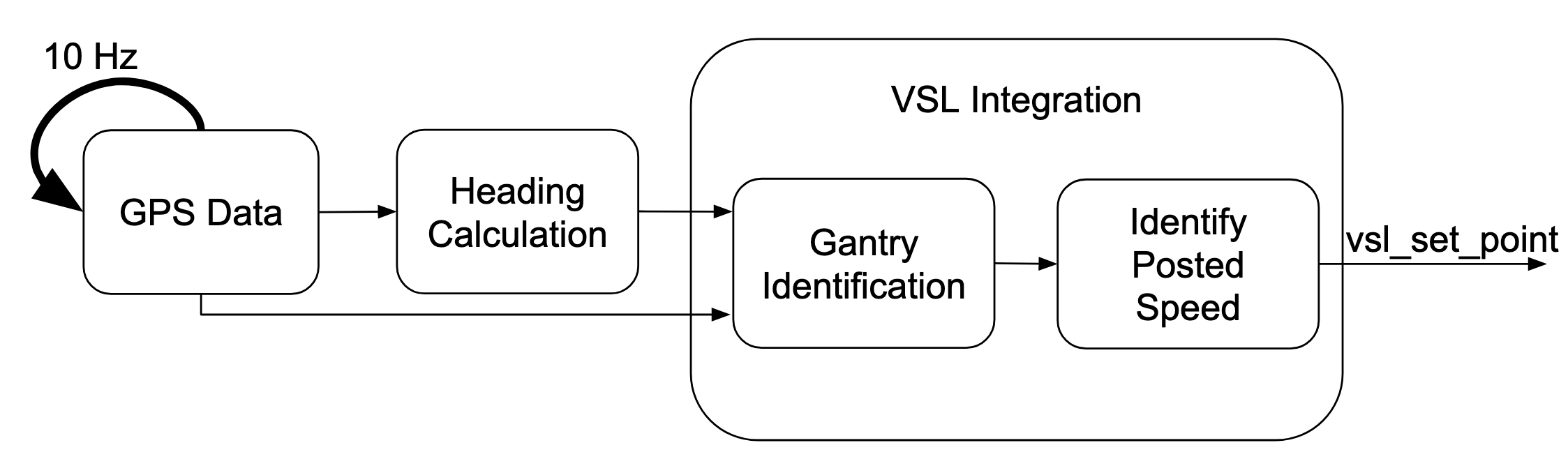}
    \caption{High level flowchart showing how our implementation starts with GPS data and produces a vehicle's desired velocity (also called a set point). See Figures~\ref{fig:gantry} and \ref{fig:setpoint} for more detail on the Gantry Identification and Identify Posted Speed components, respectively.}
    \label{fig:flowchart}
\end{figure}

\begin{figure}
    \centering
    \includegraphics[width=\linewidth]{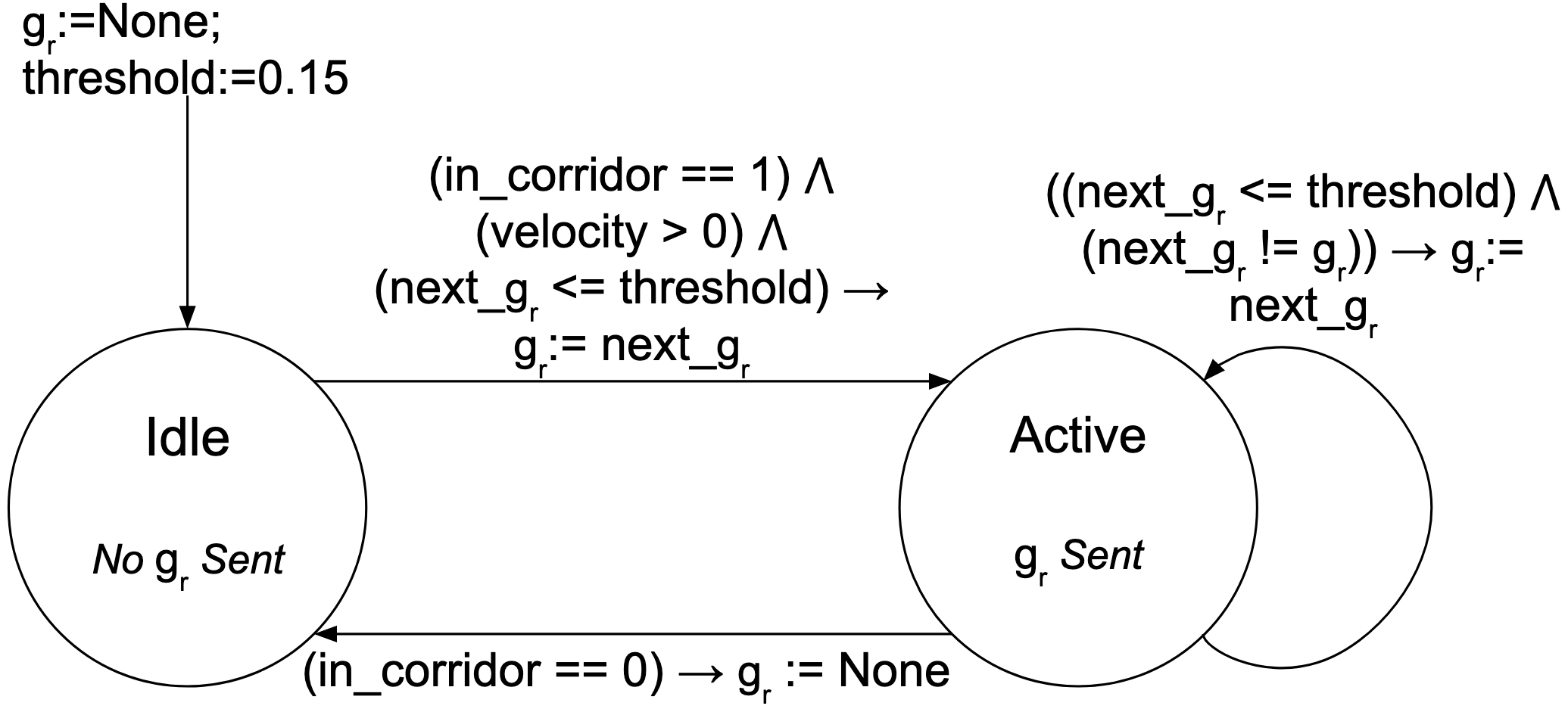}
    \caption{The Gantry Identification component has two states: Idle, where no gantry $g_{r}$ is sent forward, and Active, where a VSL gantry $g_{r}$ is sent on. When the system starts it enters as Idle. When conditions are met there is a transition to the Active state. The vehicle continues to set and hold $g_{r}$, sending it forward as well, until the vehicle leaves the SMART corridor (via freeway exit or out the ends of the instrumented freeway).}
    \label{fig:gantry}
\end{figure}

\begin{figure}
    \centering
    \includegraphics[width=\linewidth]{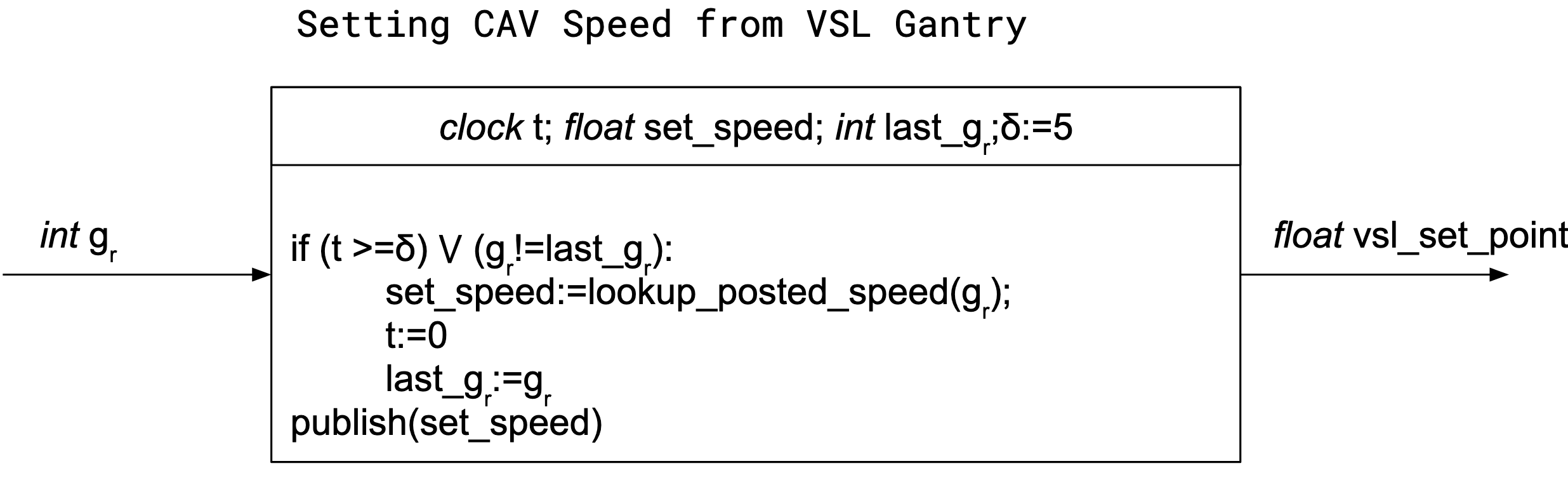}
    \caption{Identifying the $v_{g_{r}}$ takes in a sent gantry value $g_{r}$ and publishes a desired vehicle velocity in two ways. Either there has been an event where the $g_{r}$ changes, which triggers a lookup request for the posted speed, or it has been 5 seconds since the last cached posted speed has been queried so a new lookup request is triggered. This allows the vehicle to react to changes in $v_{g_{r}}$ from the same $g_{r}$, or changes in the $g_{r}$ while traveling down the freeway.}
    \label{fig:setpoint}
\end{figure}

Figure~\ref{fig:flowchart} overviews the process to finding the relevant gantry ($g_{r}$) and the VSL posted speed ($v_{g_{r}}$) for the CAV desired speed, summarized as follows. With GPS location, we can identify the location of the vehicle ($l)$ and it's heading ($h)$. Using this information, we create a simple model and algorithm to identify the relevant VSL gantry $g_{r}$ (if applicable) and identify the posted speed at the identified gantry $v_{g_{r}}$. To identify $g_{r}$, we define a polygon $p_{corridor}$ representing the set of locations where the SMART corridor lies, then proceed to calculating the state of the vehicle with respect to the VSL system (i.e. $l \subset p_{corridor}?$ and $h == westbound?$), and send forward $g_{r}$ when identified. Figure~\ref{fig:gantry} features a model showing how this functions. As implemented, the CAV considers a gantry to be $g_{r}$ when approaching it and crossing a 0.15 mile threshold. $v_{g_{r}}$ is identified with a lookup triggered either with an event (like crossing the 0.15 mile threshold) when $g_{r}$ is updated, or every 5 seconds. This allows the vehicle to react to changes in $v_{g_{r}}$ from the same gantry over time, or changes to $g_{r}$ while traveling down the freeway. Figure~\ref{fig:setpoint} summarizes this process of taking in the $g_{r}$ when sent, and publishing $v_{g_{r}}$ as the CAV desired speed. These components are tested with software-in-the-loop leveraging the default ROS\cite{quigley2009ros} playback functionality with recorded drives containing trajectories in the SMART corridor.

\subsection{Controllers}
Here we describe the different control algorithms running on the experimental vehicle throughout testing. 

Vehicular dynamics are controlled via a commanded acceleration value sent along the vehicle's CAN bus. A low-level control system implemented by the vehicular manufacturer converts this command to more specific vehicular dynamic commands (e.g. throttle,braking, engine). Let $u_{cmd}$ refer to the acceleration command send to the low-level controller.

We create values of $u_{cmd}$ through two higher level control algorithms. The first control law we refer to as the \textit{nominal controller}, which calculates acceleration commands meant to track the desired speed as per the VSL system. Let $u_{nom}$ refer to the acceleration coming from the nominal controller. Let $v$ be the measured speed of the vehicle, and $v_{g_{r}}$ be the desired speed from the relevant gantry in the VSL system. The acceleration from the nominal controller is calculated using a proportional control law of the following form:

\begin{equation}
    u_{nom} = k_{p}\left( v_{g_{r}} - v \right)
\end{equation}
where $k_p$ is the proportional gain parameter. A value of $0.8$ was used for $k_p$ in experimentation. 

In addition to the nominal controller, we use a \textit{control barrier function} (CBF) as an \textit{active safety filter} (ASF). Here we give a high-level description of the CBF used, for a more detailed description of developing CBFs as ASFs for vehicular control we direct the reader to~\cite{gunter2022experimental,gunter2022CBFs_for_cutins,ames2016control}. Let $u_{safe}$ be an acceleration command calculated using the CBF. Additionally, let $s$ be the inter-vehicle spacing, and let $v_{l}$ be speed of a preceding vehicle, both as measured by the vehicle's onboard radar system. Acceleration commands meant to supervise for safety are calculated as follows:

\begin{equation}
    u_{safe} = \frac{k_{CBF}}{t_{min}}\left(s - \left( t_{min}v + s_{min} \right) \right) + \frac{1}{t_{min}} \left( v_l - v \right)
\end{equation}
where $k_{CBF},t_{min},\text{and } s_{min}$ are control parameters which we assign values of $0.1, 2.0,$ and $15.0$ respectively. This CBF is designed to keep the vehicle's spacing-gap above a value of $t_{min}v + s_{min}$, where $t_{min}$ is a minimum time-gap, while $s_{min}$ is a minimum spacing-gap. $k_{CBF}$ is a control gain parameter. This choice of safety is common~\cite{ames2019control,ames2016control}, but not the only possible choice.

To create a union between the nominal controller and the ASF  we do the following:
\begin{equation}
    u_{cmd} = \text{min}\left(u_{nom},u_{safe} \right)
\end{equation}
which is interpretable as taking the smallest in value acceleration calculated by either the ASF or the nominal controller. This leads to overall system control which tracks velocity when safety is not a concern (not following another vehicle closely), but control that tracks for safety when needed.

\subsection{Hardware Instrumentation}
To extend a stock vehicle's hardware, we use a Raspberry Pi 4 board, a CAN interface board, a GPS module, connecting cables, and a battery-powered uninterrupted power supply. In total, the instrumentation cost comes under \$500 USD. The embedded computer is situated underneath the passenger seat, and cables to connect to the ADAS module and GPS module are routed inconspicuously to allow for a unobstructed view for the driver operator of the vehicle. Mobile phones using LTE act as a mobile hotspot to provide internet connectivity for requesting the latest VSL setpoints. 

\subsection{Experimental Design}

The experiment consisted of two vehicles which were released from a parking lot and directed to travel in the high occupancy vehicle lane (lane number 1 in the standard incident management lane numbering scheme). The speed of the first, or `pilot', vehicle was regulated by the driver, who was instructed to travel at 70 mph (the maximum speed limit) when the posted VSL messages reported 70 mph. When the VSL gantries posted speeds less than 70 mph, the driver was instructed to drive at the prevailing speed of traffic. The speed of the second vehicle, or `control vehicle', was regulated by the enhanced adaptive cruise control system developed in this work which dynamically adjusts the maximum speed of the vehicle based on the real-time variable speed limits. Both drivers were instructed to maintain a safe operating environment and to abandon the experiment if conditions on the roadway prevent a safe experiment from being executed. The experiments were conducted in the 5:30-7:30 am window in which I-24 experiences the start of morning traffic conditions and regular traffic waves develop.

\section{Results}\label{sec:results}
We characterize the performance of our control vehicle with respect to its ability to comply with the posted variable speeds, and characterize the effect of this compliance compared to the downstream pilot vehicle in the traffic flow following the prevailing traffic speeds. This analysis is achieved in part by using the \textit{strym} library\cite{bhadani2022strym}.

\subsection{Performance of CAV for VSL-Following}

\begin{figure}
    \centering
    \includegraphics[width=\linewidth]{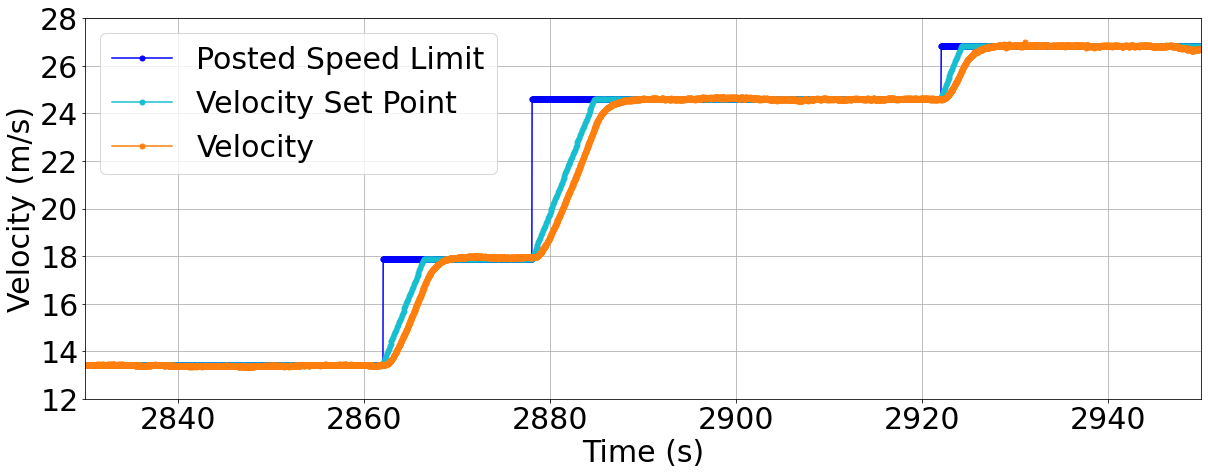}
    \caption{In the time domain, we can see the input response of the gantry, the ramping up, and then the response of the vehicle velocity.}
    \label{fig:ts1}
\end{figure}

Figure \ref{fig:ts1} shows the performance of the CAV as the posted speed limits increase. The posted speed $v_{g{_r}}$ increases (blue), the ramp filter (light blue) smooths the step up in velocity, and the measured velocity (orange) rises to remain pinned to the variable speed limit.

\begin{figure}
    \centering
    \includegraphics[width=\linewidth]{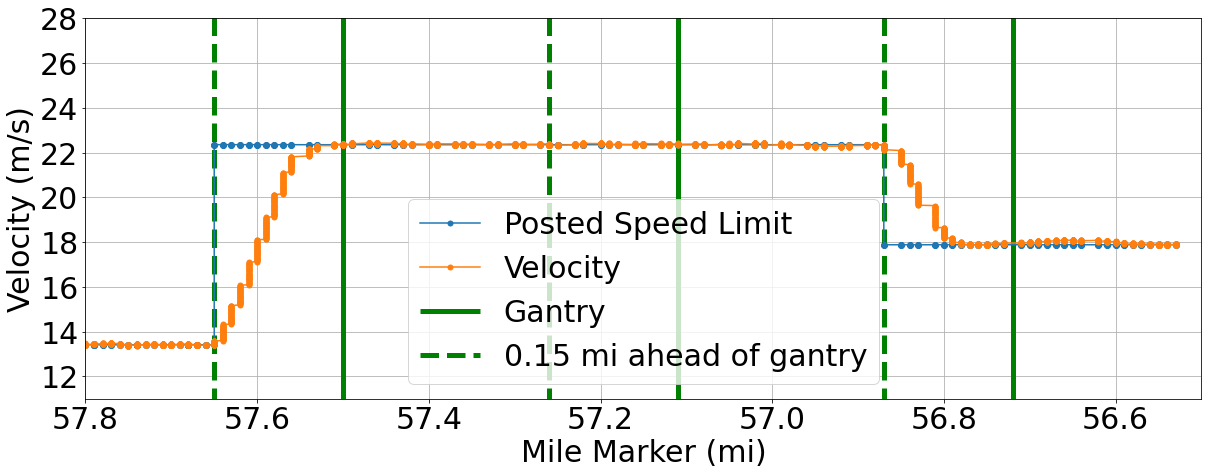}
    \caption{This is a figure showing how the vehicle is receiving new information starting at the pre-gantry dotted line, and the adjusting to the posted speed before reaching the gantry.}
    \label{fig:upAndDown}
\end{figure}

Figure~\ref{fig:upAndDown} highlights the performance of the CAV while traveling down the freeway. This trajectory segment covers the three straightforward cases when transitioning to a new gantry $g_{r}$: an increased $v_{g{_r}}$, an equal $v_{g{_r}}$, or a decreased $v_{g{_r}}$. As the CAV crosses the threshold to transition to the gantry near mile marker 57.6 $v_{g{_r}}$ is updated, and before reaching the gantry the desired velocity is reached. The posted speed limit $v_{g{_r}}$ is the same at the second gantry met in the middle of Figure~\ref{fig:upAndDown}, so we see no velocity changes. When meeting the third gantry, $v_{g{_r}}$ is read at the threshold approaching the gantry, and well before passing the sign the CAV is travelling at the new lower $v_{g{_r}}$.

\begin{figure}
    \centering
    \includegraphics[width=\linewidth]{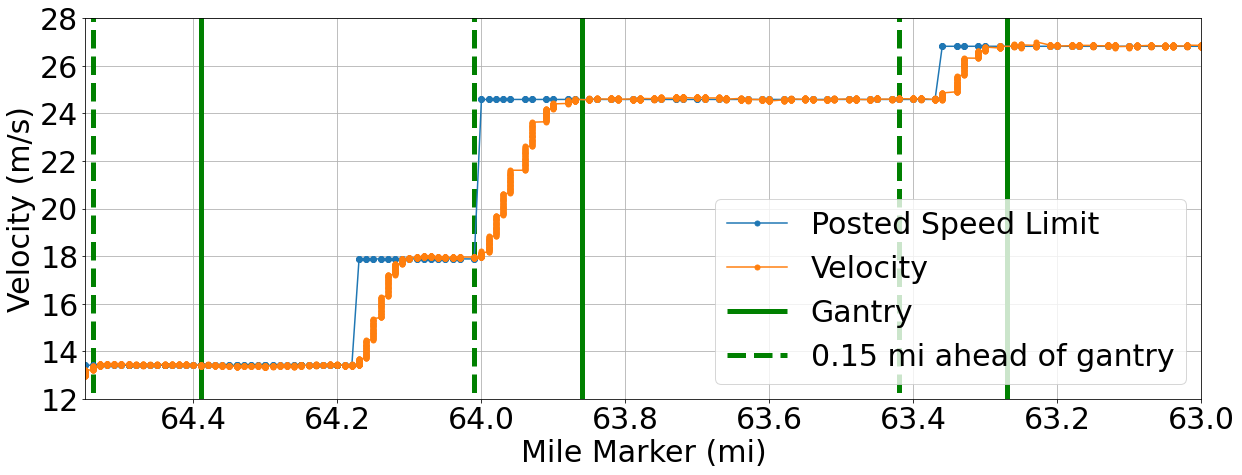}
    \caption{This figure shows further than Figure \ref{fig:upAndDown}, that our implementation is a 'set and hold' at each passing gantry. The gantry near mm 64.4 increases the posted speed to 18 $\frac{m}{s}$ before we reach the next gantry, so we increase our speed. When reaching the next 'update zone' for gantry at mm 63.85, we update our velocity setting again.}
    \label{fig:stepup}
\end{figure}

The trajectory segment in Figure~\ref{fig:stepup} is the same as in Figure~\ref{fig:ts1} but new insights are revealed when plotting in the roadway coordinate domain. There are three steps up in velocity in this trajectory. Recall, we choose an implementation which sets-and-holds the gantry $g_{r}$, while continuing to listen to changes in the posted speed $v_{g{_r}}$ which may occur until switching to the next gantry. The first step up past mile marker 64.2 reflects a change in the $v_{g{_r}}$ from the $g_{r}$ passed by earlier at mile marker 64.4. The second step up in velocity occurs in a straightforward manner -- when crossing the threshold to the next gantry. Note that again the $v_{g{_r}}$ is reached before passing under the gantry. The third step up shows that as we cross the threshold to the gantry near mile marker 63.2, $v_{g{_r}}$ has not changed. As the control vehicle continues to approach the $g_{r}$, the $v_{g{_r}}$ and the CAV responds accordingly by increasing its velocity.

\begin{figure}
    \centering
    \includegraphics[width=\linewidth]{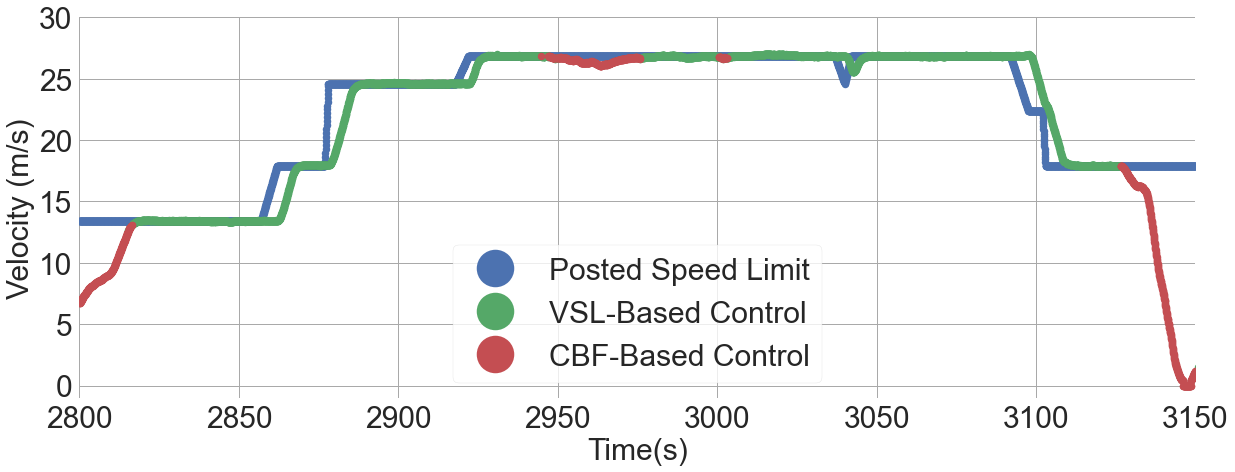}
    \caption{Multiplexed control states: while satisfying forward collision avoidance (control barrier), the vehicle software makes best efforts to match the posted speed limit and remain comfortable to ride in.}
    \label{fig:cbf-active}
\end{figure}

The control vehicle is not always able to cleanly follow velocity limits as posted by the VSL system. A heavily congested interstate highway is a complex environment in which safe car-following often takes priority over the ability to travel at the speed limit. Figure~\ref{fig:cbf-active} features the mode switching in our multiplexed control system, which allows for smooth transitions between variable speed limit following and keeping track of safety. 

The control vehicle should have a reasonable rise/fall time (interval from when a new $v_{g_{r}}$ is ingested and the vehicle reaches $v_{g_{r}}$). Due to the small-scale of this deployment and the complex nature of the wild congested freeway environment, we report simple statistics on these events to claim generally reasonable behavior. There were 8 rising events in our experimental drives prompted from VSL system, ranging from a minimum rise time of 3.97 seconds to a maximum of 11.50 seconds, and with a mean of 6.21 seconds. The changes in $v_{g_{r}}$ range from $2.24 \frac{m}{s}$ to $8.9 \frac{m}{s}$. There were 6 falling events, ranging from a minimum fall time of 5.21 seconds to a maximum of 8.08 seconds, and a mean of 6.79 seconds. The changes in $v_{g_{r}}$ range from $-2.24 \frac{m}{s}$ to $-4.47 \frac{m}{s}$. This behavior is in line with the system design and implementation, and results in comfortable and reasonably prompt changes to the vehicle's velocity as dictated by the VSL system.


\subsection{Potential Benefits}

\begin{figure}
    \centering
    \includegraphics[width=\linewidth]{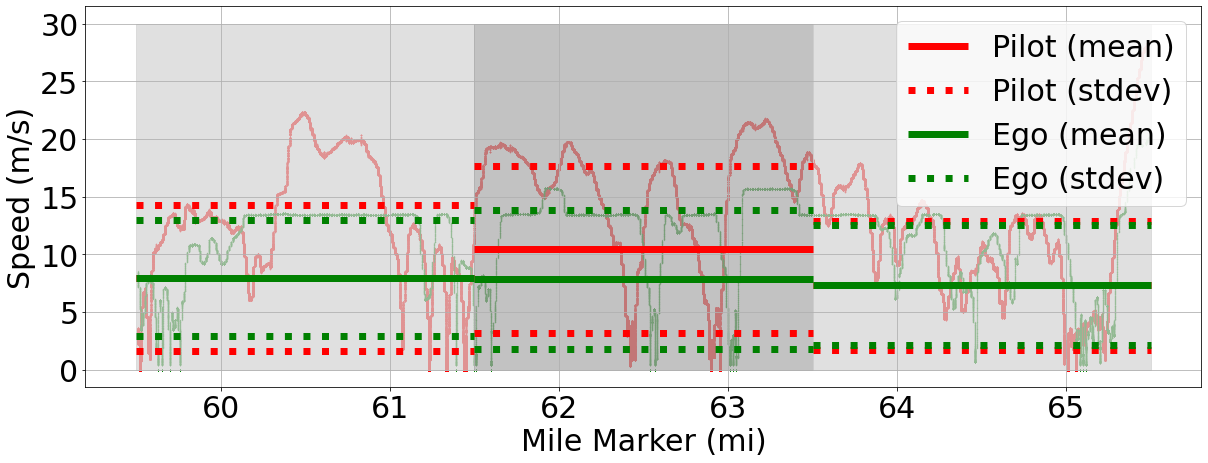}
    \caption{Less deviation is shown in the ego car than the non-controlled car, but average speed is very similar. Ego behavior is preferred when it comes to traffic smoothing, and increasing energy efficiency.}
    \label{fig:v-stdev}
\end{figure}

The lead (pilot) car and the control car (ego) left at nearly identical times, and drove in the same lane. Their speed plotted by the section of the roadway is shown in Figure~\ref{fig:v-stdev}, with overlay of the mean speed and standard deviation divided into three segments. As the standard deviation is normalized by the number of samples, it is roughly equivalent to the \textit{normalized mean-squared error} (NMSE).

It is noteworthy that the average speed over this time is very similar in the first and last segment, with a larger average speed in the middle segment for the pilot car. However, the standard deviation of the ego vehicle is between 10\%-25\% lower than the standard deviation of the pilot vehicle, compared to its average speed in those areas. This is consistent with reductions found in \cite{stern2018dissipation}, which indicates that there could be significant energy and safety benefits, and that there is motivation for additional study.

\begin{table}[]
    \centering
\begin{tabular}{|l|c|c|c|c|}%
\hline
\bfseries mm & \textbf{Pilot (NMSE/Mean)} & \textbf{Ego (NMSE/Mean)}
\csvreader[head to column names]{speeds.csv}{}
{ \\\hline\mm & (\pilotmse/\pilotmean) & (\egomse/\egomean)}
\\\hline
\end{tabular}
    \caption{Average speed and MSE for each car shown in Fig.~\ref{fig:v-stdev}.}
    \label{tab:norms}
\end{table}





\subsection{Practical Findings for Future Deployments}
There are psychological and safety environment components which are tested by complying with the $v_{g_{r}}$. Nearby drivers are frustrated by going much slower than prevailing conditions, even though they are speeding into stopped traffic. It is more challenging to comply with $v_{g_{r}}$ in an environment where prevailing speeds are greater than $10 \frac{m}{s}$ faster, as we measured.

\section{Conclusions and Future Work}\label{sec:conclusions}

Our work demonstrates it is possible to deploy CAVs with the capability to follow publicly broadcast variable speed limits at scale with LTE connectivity and affordable hardware. The field experiment shows strong performance of ADAS-equipped vehicles that react through connectivity to Variable Speed Limit systems in the field. The vehicle's reaction is correct with respect to specification within the geographical range (zone) of each VSL gantry, and responds as designed when a posted speed limit changes while moving within a gantry zone. 
The comparison between the pilot vehicle and control vehicles show nearly identical travel times and average speeds, but the variation in speed of the pilot vehicle is notably higher than the ADAS-equipped control vehicle.

The results go beyond simply driving at the posted speed limit and presenting the driving data. The impact includes an explicit cost that could enable this adoption at scale: less than \$500 per vehicle, using connectivity over existing LTE networks to a tethered mobile phone without requiring 5G or high-speed radios. 

This article demonstrated the feasibility and efficacy of the CAV platform. The traffic and safety analysis for this small-sized test show promising indicators, and motivate future work to scale the experiment to a larger number of vehicles in order to evaluate the efficacy of the system deployment on many vehicles simultaniously. 



\section*{Acknowledgment}

This work is supported by the National Science Foundation under awards 2111688 and 2135579, and the Dwight D. Eisenhower Fellowship program under Grant No. 693JJ32345023. George Gunter was supported through an NSF Graduate Research Fellowship. The authors would like to thank the Tennessee Department of Transportation, Southwest Research Institute, Stantec, and Arcadis, who manage and operate the variable speed limit system for the I-24 SMART Corridor, and Yuhang Zhang and Junyi Ji from Vanderbilt University for their support on vehicle testing.

\clearpage
\bibliographystyle{IEEEtran}
\bibliography{references}

\end{document}